%% file: main.tex
\documentclass{article}

\PassOptionsToPackage{numbers,compress}{natbib}
\usepackage[preprint]{neurips_2026}

\usepackage[utf8]{inputenc} % allow utf-8 input
\usepackage[T1]{fontenc}    % use 8-bit T1 fonts
\usepackage{hyperref}       % hyperlinks
\usepackage{url}            % simple URL typesetting
\usepackage{booktabs}       % professional-quality tables
\usepackage{amsfonts}       % blackboard math symbols
\usepackage{nicefrac}       % compact symbols for 1/2, etc.
\usepackage{microtype}      % microtypography
\usepackage{xcolor}         % colors

\usepackage{graphicx}
\usepackage{tabularx}
\usepackage{multirow}
\usepackage{wrapfig}
\usepackage{subcaption}
\usepackage{caption}
\usepackage{array}
\usepackage{enumitem}
\usepackage{pifont}  % for \ding{51} (checkmark) and \ding{55} (crossmark)
\newcolumntype{Y}{>{\raggedright\arraybackslash}X}

\usepackage{amsmath}
\usepackage{amssymb}

\title{Explicit Critic Guidance for Aligning Diffusion Models}

\author{%
  Zhengyang Liang \\
  University of Toronto \\ Vector Institute \\
  \And
  Qihang Zhang \\
  The Chinese University of Hong Kong \\
  \And
  Ceyuan Yang \\
  The Chinese University of Hong Kong \\
}

\begin{document}

\maketitle

\input{section/abstract}

\input{section/intro}
\input{section/related}
\input{section/preliminary}
\input{section/method}

\input{section/exp}

\section{Conclusion}
We present a stable, efficient, and scalable actor–critic framework for post-training diffusion models. Our method makes three contributions: (1) converting diffusion backbones into timestep-aware critics with value pretraining enables stable and sample-efficient PPO training; (2) a lightweight multi-value-head design supports multi-reward optimization; and (3) the learned critic can be reused for inference-time steering, providing performance gains beyond training. These components establish a practical recipe for applying reinforcement learning to modern diffusion models.

% ---- Bibliography ----
% NeurIPS uses natbib (loaded by neurips_2026.sty)
% Use plainnat style for natbib compatibility
\bibliographystyle{plainnat}
\bibliography{main}

%%%%%%%%%%%%%%%%%%%%%%%%%%%%%%%%%%%%%%%%%%%%%%%%%%%%%%%%%%%%

\appendix
\input{section/X_syppl}

%%%%%%%%%%%%%%%%%%%%%%%%%%%%%%%%%%%%%%%%%%%%%%%%%%%%%%%%%%%%

\end{document}

%% file: section/abstract.tex
\begin{abstract}
Online reinforcement learning is becoming increasingly important for aligning diffusion models with non-differentiable objectives. However, existing methods still face limitations in assigning fine-grained credit along denoising trajectories and in realizing stable value-based optimization. We propose a state-aligned latent actor-critic framework for diffusion post-training, in which the diffusion model serves as its own timestep-conditioned value function and predicts values directly on noisy latent states. This enables trajectory-level PPO training, supports stable actor-critic optimization with simple conditioning and value pretraining strategies, and naturally allows the learned critic to be reused for inference-time steering. We further extend the framework to multi-reward optimization, where joint training with complementary rewards helps alleviate reward hacking. Across both UNet- and DiT-based backbones, our method consistently outperforms prior group-relative RL and actor-critic baselines on single-reward and multi-reward benchmarks, while test-time steering provides additional gains in generation quality.
\end{abstract}

%% file: section/intro.tex
\section{Introduction}
\label{sec:intro}

Post-training is becoming an increasingly important stage for visual generative models. Although large-scale pretraining~\cite{esser2024scaling, seawead2025seaweed, wu2025qwen, liang2024scaling, yin2025towards} already yields strong visual quality, pretrained models often remain suboptimal on properties that users directly care about, including prompt faithfulness, compositional correctness~\cite{ghosh2023geneval, huang2023t2icompbench}, text rendering~\cite{cui2025paddleocr3}, aesthetics, and human preference~\cite{wu2023human, ma2025hpsv3, xu2023imagereward, kirstain2023pick, wang2025unifiedreward}. Reinforcement learning~\cite{schulman2017proximal, ahmadian2024back, li2023remax, shao2024deepseekmath, zheng2025group, schulman2015trust, hu2025reinforce++, ouyang2022training, stiennon2020learning, bai2022training} is particularly appealing in this setting because it enables optimization with respect to verifiable and task-specific objectives, including learned preference models and rule-based rewards.

\begin{figure*}[!t]
    \centering
    \includegraphics[width=\linewidth]{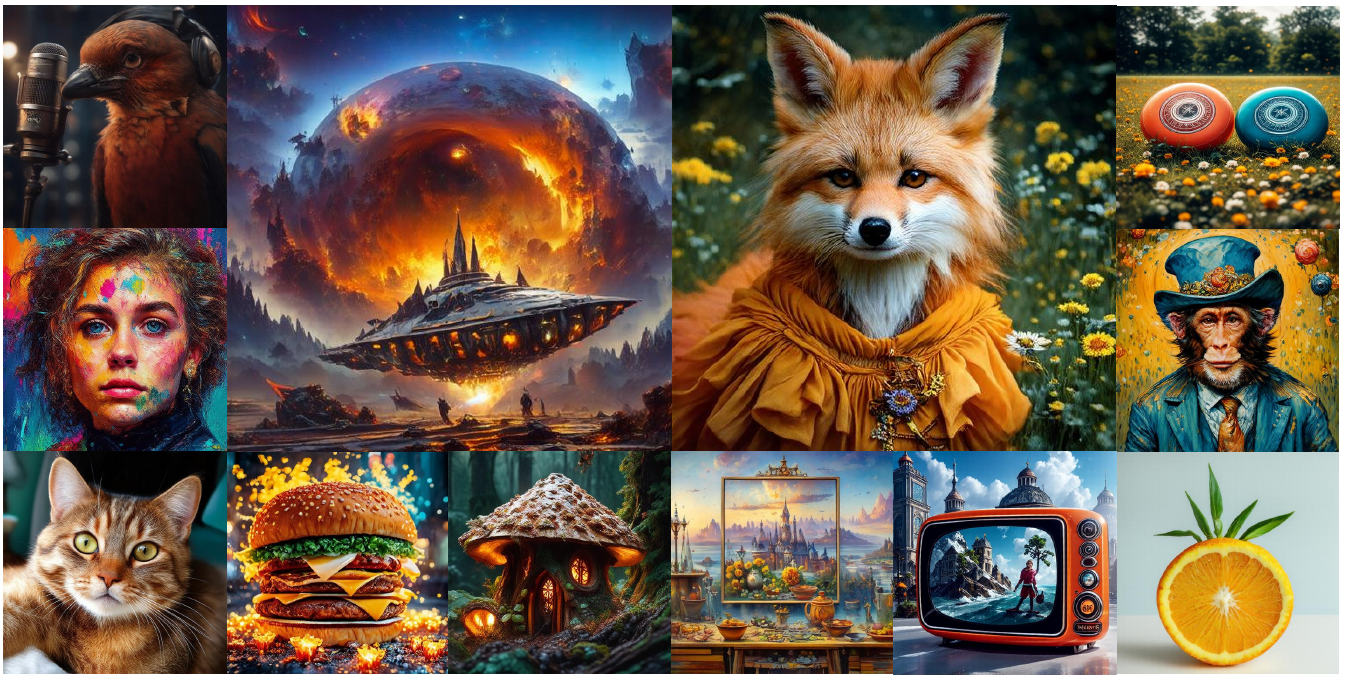}
    \caption{\textbf{Text-conditioned image generation results from our SD3.5M models.} All samples are generated \textbf{without} classifier free guidance.}
    \label{fig:teaser}
\end{figure*}

A central challenge in diffusion post-training is \emph{credit assignment}. Diffusion models~\cite{ho2020denoising, song2019generative, song2020score, dhariwal2021diffusion, lipman2023flow, liu2023flow} generate samples through multi-step denoising trajectories, whereas rewards are typically defined only on the final decoded image. The problem is therefore not only to maximize terminal reward, but to assign it to the intermediate noisy states and denoising decisions that produced it. Group-relative methods~\cite{liu2025flowgrpo, xue2025dancegrpo, deng2026densegrpo} partially avoid this issue by comparing samples within a group, but the resulting signal remains coarse and trajectory-level. Explicit actor-critic methods are in principle better suited to this setting, yet existing diffusion critics~\cite{zhao2025score, mcallister2025flowmatchingpolicygradients} are typically defined in pixel space: they evaluate reconstructed clean-image proxies rather than the actual noisy latent states visited by the policy. This mismatch weakens alignment between critic and policy, makes value estimation brittle for structured or discrete rewards, and incurs substantial decoding overhead. Moreover, explicit actor-critic training is often unstable, as poorly conditioned value estimates can quickly undermine policy optimization.

To address these issues, we propose a \emph{state-aligned latent actor-critic} framework in which the diffusion model itself serves as a timestep-conditioned value network and predicts $V_\phi(z_t,t,y)$ directly on noisy latent states. This requires only lightweight modifications to the pretrained diffusion backbone, allowing the critic to inherit the model's generative prior while operating on the same state space and diffusion-native features as the policy. The resulting critic is naturally noise-aware, avoids repeated image decoding, and can be reused at inference time, where its latent-space gradients provide a direct steering signal during sampling. To make this formulation practical, we further identify timestep-aware conditioning and a short value pretraining stage as simple but effective techniques for improving training stability and efficiency. As illustrated in Fig.~\ref{fig:teaser}, the framework also enables \emph{CFG-free} post-training, reducing memory overhead and training cost while avoiding the color saturation artifacts and instability often introduced by classifier-free guidance.

A stronger optimizer, however, also makes reward misspecification more visible. In diffusion post-training, reward hacking often appears as shortcut solutions that improve the target metric while degrading unmeasured aspects of image quality. For example, optimizing a compositional reward in isolation may improve the target score while sacrificing background diversity or aesthetics, whereas complementary rewards can counteract such collapse modes. This motivates our multi-reward extension: we equip the shared latent backbone with separate value heads so that multiple rewards can regularize one another while still benefiting from a common timestep-aware representation. In this way, multi-reward optimization provides a practical mechanism for mitigating proxy overoptimization in diffusion RL.

In summary, our contributions are as follows:
\begin{itemize}[leftmargin=*,itemsep=1pt,parsep=0pt,topsep=2pt]
\item We propose a state-aligned latent critic framework for diffusion post-training, which converts a pretrained diffusion model into its own value network with lightweight modifications and supports both latent-state value prediction and inference-time steering.
\item We identify timestep-aware conditioning and value pretraining as simple but effective techniques for stabilizing and accelerating explicit actor-critic diffusion training.
\item We extend the framework to multi-reward optimization, where joint training with complementary rewards helps mitigate proxy overoptimization and alleviate reward hacking.
\item Across both single-reward and multi-reward benchmarks, our method consistently outperforms prior group-relative RL and actor-critic baselines, with additional gains from inference-time steering.
\end{itemize}

%% file: section/related.tex
\section{Related Work}
\label{sec:related}

\noindent\textbf{Diffusion alignment.}
Recent work on aligning diffusion models can be broadly grouped into reward-based fine-tuning, preference optimization, and online reinforcement learning. Early approaches align text-to-image models with human or learned feedback by collecting preference data, training reward models, and fine-tuning the generator with reward-based objectives~\cite{lee2023aligning, fan2025online, peng2019advantage, xue2025advantage}. For differentiable rewards, alignment can also be performed by backpropagating reward gradients through the denoising process~\cite{xu2023imagereward, clark2024directly, prabhudesai2023aligning, prabhudesai2024videodiffusionalignmentreward}. Another line of work adapts preference optimization methods to diffusion models, including DPO-style formulations~\cite{rafailov2023direct, wu2025densedpo, liu2025videodpo, liang2025aesthetic, li2024aligning, zhang2025diffusion,zhang2024onlinevpo, furuta2024improving, yang2024using}. More recently, online RL methods~\cite{black2024training, fan2023dpok, schulman2017proximal, gupta2025simple, miao2024training, zhao2025score, mcallister2025flowmatchingpolicygradients, zhang2024large} optimize diffusion models directly against downstream rewards. This line has also been extended to Group-Relative style policy gradient method~\cite{liu2025flowgrpo, xue2025dancegrpo, li2025branchgrpo, li2025mixgrpo}, which bypass the training of critic. Compared with these methods, our work focuses on value-based post-training with an explicit critic learned on the same noisy latent trajectory as the policy. We provide extended discussion in Appendix~\ref{app:ext_related}.

%% file: section/method.tex
\section{Method}
\label{sec:method}

\subsection{Problem setup}

\begin{figure*}[!t]
    \centering
    \includegraphics[width=\linewidth]{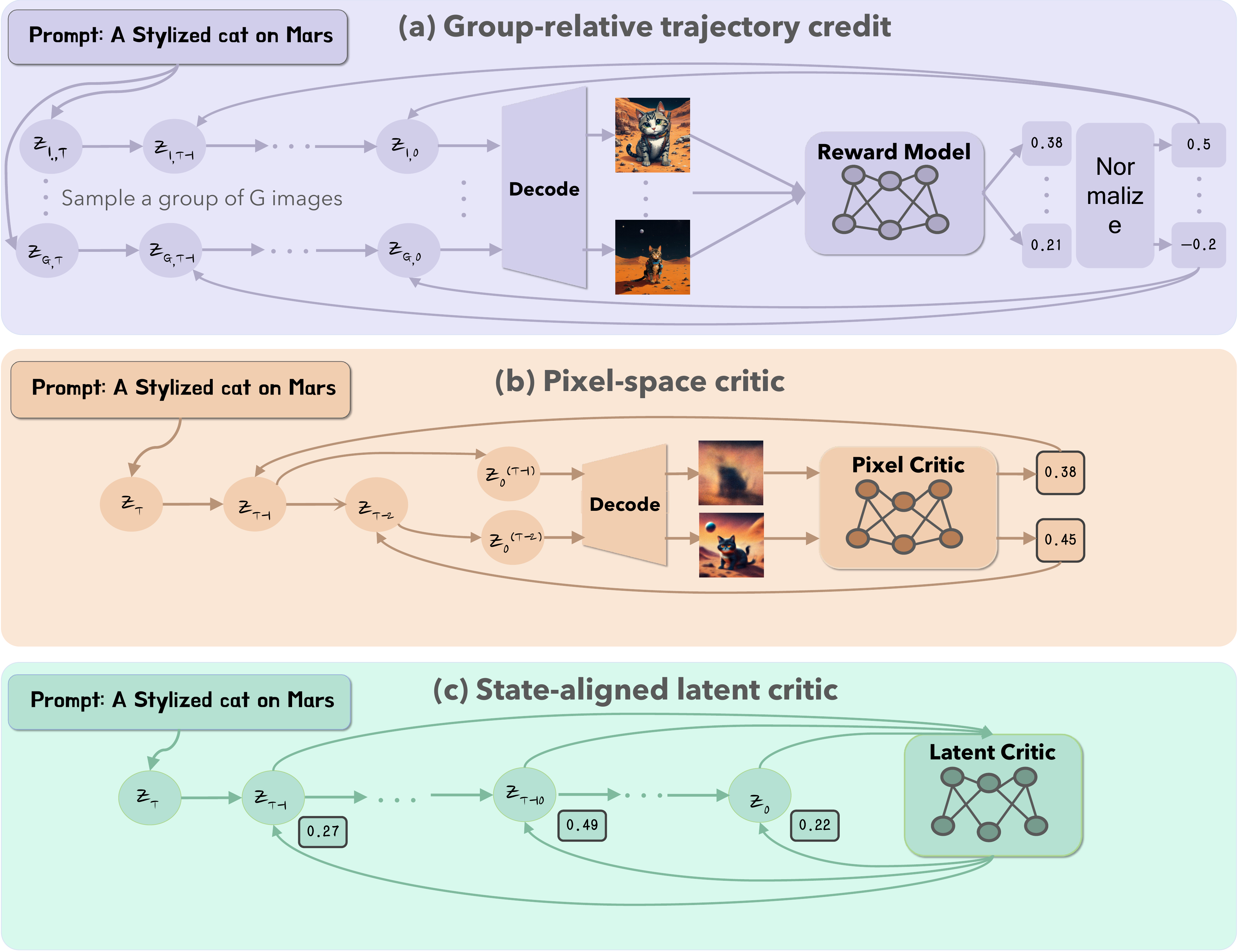}
    \caption{\textbf{Illustration of three credit assignment strategies in diffusion RL.}
GRPO-style methods assign one relative signal to each trajectory by groupwise reward comparison.
Pixel-space critics score reconstructed images decoded from intermediate latents, leading to proxy-state evaluation.
Our method instead learns a critic directly on noisy latent states, enabling state-dependent and state-aligned credit assignment as well as inference-time steering from the same learned critic.}
    \label{fig:credit}
\end{figure*}

We briefly define the sequential decision-making view of diffusion post-training used throughout the paper. Let $z_t$ denote the noisy latent at timestep $t$, and let $y$ be the text condition. A latent diffusion or flow-based generator produces an image by iteratively updating $z_t$ along a denoising trajectory from noise to data. We view this process as a finite-horizon sequential decision problem. At step $k$, the state is
\[
s_k = (z_{t_k}, t_k, y),
\]
and the action $a_k$ is the model prediction used by the sampler, e.g., a noise, score, or velocity prediction depending on the backbone. The transition is induced by the sampler. After the final step, the denoised latent is decoded and evaluated by a reward function, yielding a terminal reward $r_T = r(x, y)$, with $r_k = 0$ for $k < T$. The objective is to maximize the expected terminal reward over the denoising trajectory. To obtain intermediate feedback, we learn a timestep-conditioned value function
\begin{equation}
V_\phi(s_t, t) \;\approx\; \mathbb{E}_{\pi}\!\left[\, R \mid s_t \,\right],
\label{eq:value}
\end{equation}
which estimates the expected terminal return from an intermediate noisy latent state. The critic is then used to compute advantages for PPO-style policy updates. We optimize the actor and critic network following \cite{schulman2017proximal}.
\subsection{Learning a State-Aligned Critic in Latent Space}

\begin{figure}[t]
    \centering
     \includegraphics[width=\linewidth]{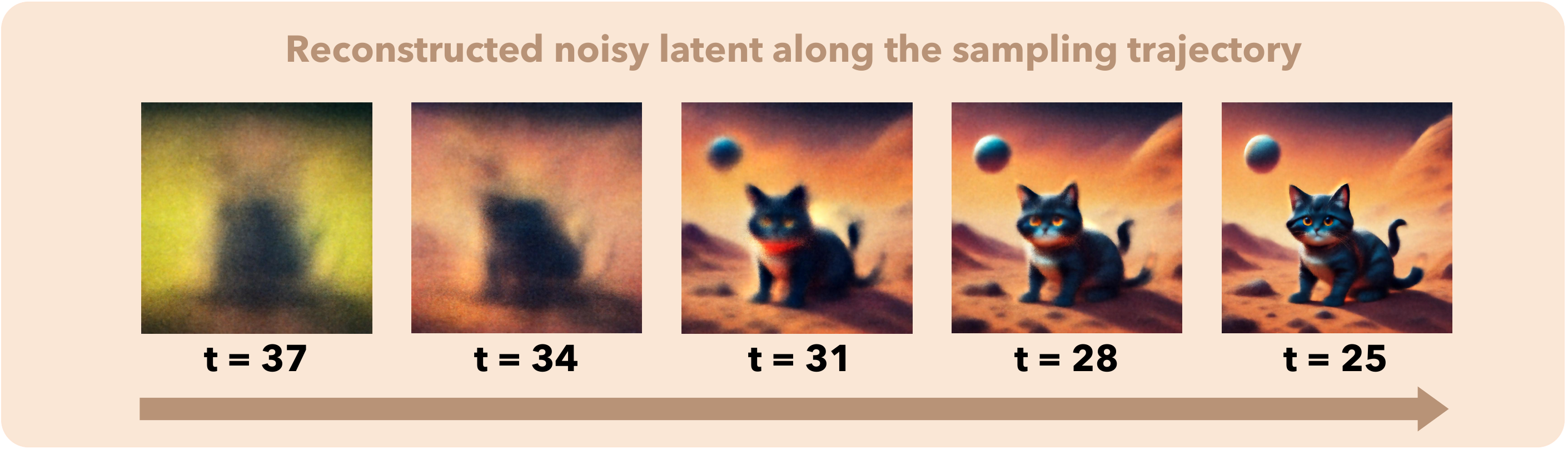}
    \caption{\textbf{Reconstructed noisy latent along the sampling chain.} We visualize selected intermediate latent states by reconstructing them into image space. Although early reconstructions are blurry and out of distribution for pixel-space evaluators, the underlying noisy latents remain on the diffusion denoising trajectory, motivating a critic that operates directly in latent space.}
    \label{fig:traj}
\end{figure}

\noindent\textbf{Credit assignment.}
As discussed in Sec. 1, existing work either bypasses explicit critic or learning diffusion critics operate on proxy states rather than the actual noisy latents visited by the policy. We address this by learning a critic directly on the denoising trajectory. We provide an illustration of three mechanisms in Fig.~\ref{fig:credit}. Moreover, as we show in Fig.~\ref{fig:traj}, such reconstructed images are often out of distribution for pixel-space evaluators such as CLIP~\cite{radford2021learning} or BLIP~\cite{li2022blip}, whereas the corresponding noisy latents remain in distribution for the diffusion model itself. Pixel critic requires extra finetuning to adapt to these inputs. These observations motivate our central design choice: effective diffusion RL requires a critic that assigns value directly on the noisy latent trajectory.

\begin{figure*}[h]
    \centering
     \includegraphics[width=\linewidth]{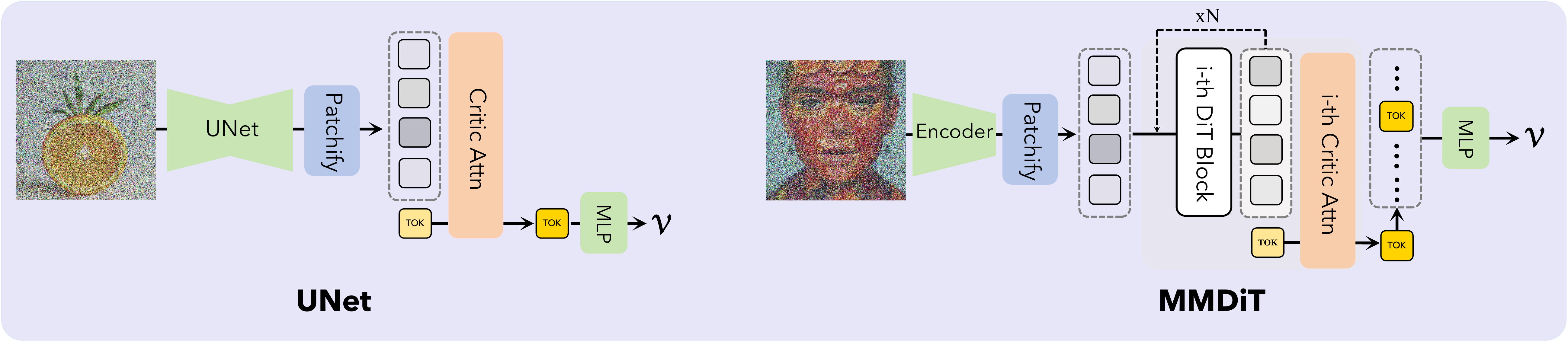}
    \caption{\textbf{Architecture overview over UNet and Diffusion Transformer.} We initialize the value network with pretrained diffusion models. Attention layer and MLP are added to predict critic scalar.}
    \label{fig:arch}
\end{figure*}

\noindent\textbf{State-aligned Latent Critic} Based on this view, we learn a timestep-conditioned latent critic
\[
V_\phi(z_t, t, y),
\]
directly on the denoising trajectory. Unlike pixel-space critics, our critic stays entirely in latent space, shares the same trajectory representation as the policy, and remains naturally aware of timestep and noise level. We initialize the critic from the actor, which is a pretrained diffusion model, and convert it into a scalar value predictor with lightweight architectural modifications. The overview for our architecture design is shown in Fig.~\ref{fig:arch}.

\noindent\textbf{UNet critic.}
For a UNet backbone, we first process the noisy latent with the standard denoising network, then patchify the resulting feature map into tokens. A learnable summary token is appended and passed through a lightweight attention head, followed by a small MLP that predicts the scalar value $V_\phi(z_t,t,y)$. 

\noindent\textbf{DiT critic.}
For a diffusion transformer~\cite{peebles2023scalable}, inspired by previous work~\cite{lin2025diffusion}, we insert lightweight critic branches at a few intermediate (layer 7, 15, 23) layers. Each branch uses a learnable query token to aggregate information from image tokens, producing a critic token for that layer. The critic tokens from selected layers are concatenated and mapped to a scalar value through an MLP head. 

\noindent\textbf{Stabilizing Critic Training.}
Directly training the latent critic is often unstable and inefficient. In practice, the critic is initialized from a pretrained diffusion backbone whose original output is a tensor-valued denoising prediction rather than a scalar return, which leads to poorly scaled value estimates and noisy policy updates. To improve critic learning, we adopt two simple techniques:
(1)~\textbf{Timestep conditioning on the value head.} Although the shared backbone already receives $t$ as input, we find it beneficial to additionally condition the final value head on the timestep via AdaLN-style modulation. This gives the head direct access to the noise level, which improves value prediction across the denoising trajectory.
(2)~\textbf{Value pretraining.} Before joint actor-critic optimization, we run a short warm-up stage in which trajectories are sampled from the current policy but only the critic is updated. Concretely, given an intermediate state $s_k=(z_{t_k}, t_k, y)$ and its Monte Carlo return target $\hat{R}_k$, we optimize the critic with a regression loss
\begin{equation}
    \mathcal{L}_{\mathrm{value}}(\phi)
    =
    \mathbb{E}_{(s_k,\hat{R}_k)}
    \left[
    \bigl(V_\phi(s_k)-\hat{R}_k\bigr)^2
    \right].
    \label{eq:value_pretrain}
\end{equation}
This warm-up yields better value estimates and improves the stability and efficiency of subsequent PPO training.

\noindent\textbf{Inference-time steering.}
Because our critic is defined on the same noisy latent states as the policy, it can be directly reused at inference time as a guidance module~\cite{ho2022classifierfreediffusionguidance, dhariwal2021diffusion}. Intuitively, the gradient $\nabla_{z_t} V_\phi$ points toward latent regions where the critic predicts higher terminal reward, playing a role analogous to classifier guidance but with reward replacing class likelihood. We compute the steering direction as:
\begin{equation}
g_t \;=\; \nabla_{z_t} V_\phi(z_t,t,y).
\label{eq:critic_grad}
\end{equation}
We then inject this direction into the sampler update. For a DDIM~\cite{song2021denoising}-style sampler with noise prediction $\epsilon_\theta(z_t,t,y)$, we modify the predicted noise as
\begin{equation}
\hat{\epsilon}_\theta(z_t,t,y)
\;=\;
\epsilon_\theta(z_t,t,y)\;-\;\eta\,\sqrt{1-\bar{\alpha}_t}\, g_t,
\label{eq:ddim_critic_guidance}
\end{equation}
where $\eta$ controls the steering strength and the factor $\sqrt{1-\bar{\alpha}_t}$ matches the perturbation magnitude to the noise level at timestep $t$.
For flow-matching / rectified-flow~\cite{liu2023flow, lipman2023flow} samplers with velocity prediction $v_\theta(z_t,t,y)$, we add the critic direction directly to the predicted velocity:
\begin{equation}
\hat{v}_\theta(z_t,t,y)
\;=\;
v_\theta(z_t,t,y)\;+\;\eta\, g_t.
\label{eq:flow_critic_guidance}
\end{equation}
In both cases, the critic-derived direction biases the denoising trajectory toward regions the critic predicts as higher-value, effectively turning the PPO-trained critic into an inference-time guidance module.

\subsection{Multi-Reward Optimization to mitigate reward hacking}

\begin{figure*}[!ht]
    \centering
    \includegraphics[width=\linewidth]{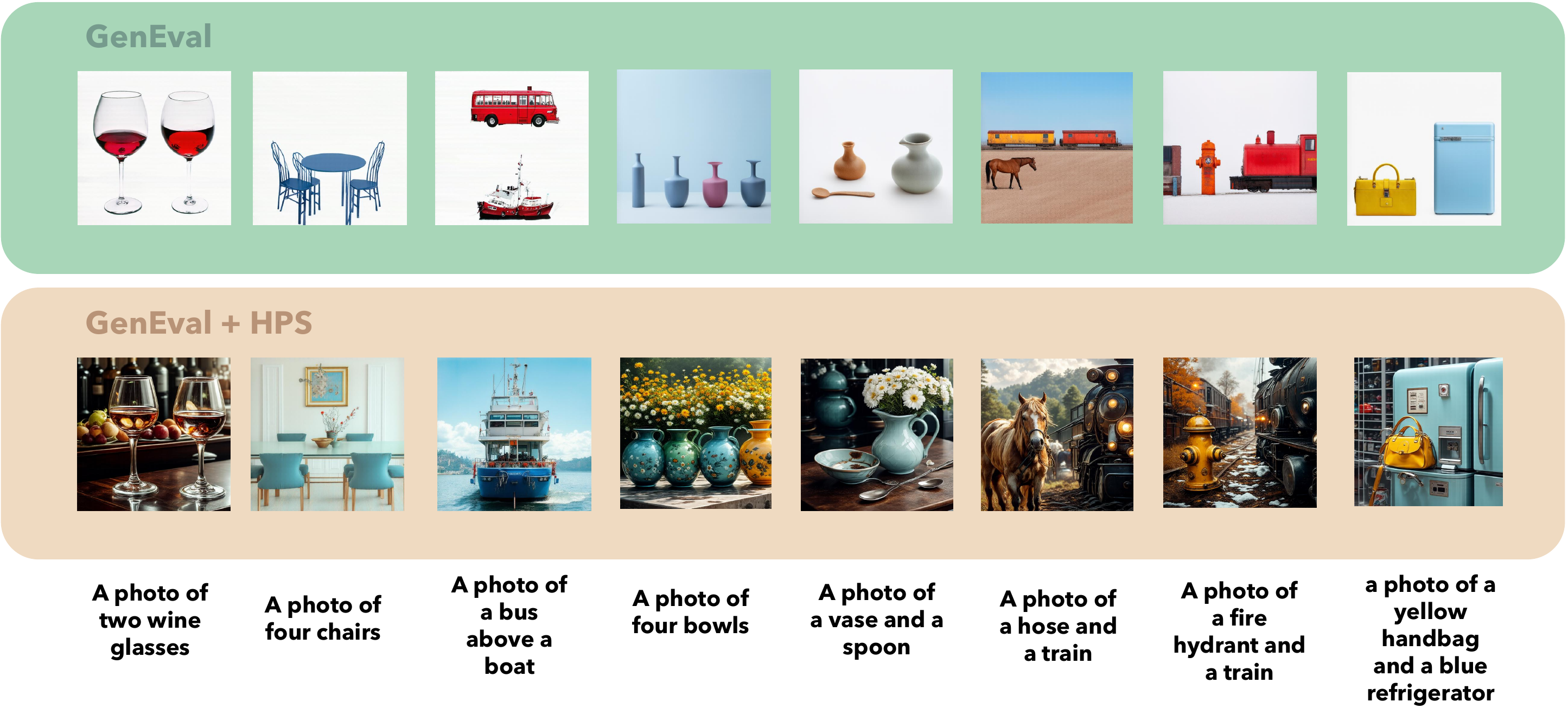}
    \caption{\textbf{Qualitative comparison of single-reward and multi-reward optimization.} GenEval-only training (top) often produces shortcut solutions with simplified composition and missing background content. Adding HPS (bottom) preserves the target objects while improving overall visual quality.}
    \label{fig:multi_reward}
\end{figure*}

\noindent\textbf{Reward hacking.}
We observe a consistent form of reward hacking during diffusion post-training: the model improves the target reward by exploiting shortcut patterns that degrade unmeasured aspects of image quality. For human-preference rewards, hacking often manifests as collapsing to recurring styles, colors, or visual templates favored by the reward model; for verifiable rewards, it is typically more severe, e.g., missing or simplified backgrounds and repetitive reward-favored structures (Fig.~\ref{fig:multi_reward} (top)). A common mitigation is KL regularization toward the base model, which indeed slows reward growth by pulling the policy back toward the pretrained distribution. However, we find that KL primarily changes the optimization timescale: when KL-regularized and non-regularized runs reach comparable reward values, they exhibit qualitatively similar hacking behavior, indicating that KL alone does not resolve reward misspecification.

% \noindent\textbf{Multi-reward as a practical mitigation.}
% Reward hacking is especially pronounced when optimizing verifiable rewards, and can emerge very early in training. As illustrated in Fig.~\ref{fig:multi_reward}, optimizing GenEval alone often drives the model toward shortcut solutions with simplified composition and missing background content, even when the target objects are correctly generated. Adding an auxiliary human-preference reward, such as HPS provides an effective counterbalance: the resulting samples retain the desired objects while exhibiting richer scenes, better aesthetics, and more natural compositions. In this sense, multi-reward training mitigates the most obvious collapse modes and delays the onset of reward hacking. However, it does not eliminate the problem. With sufficiently long training, shortcut behaviors can still dominate, so in our current setting, early stopping remains the most reliable practical safeguard, while multi-reward optimization mainly reduces the severity and slows the progression of reward hacking.
\noindent\textbf{Multi-reward optimization.}
Reward hacking is especially pronounced when optimizing verifiable rewards, and can emerge very early in training. As illustrated in Fig.~\ref{fig:multi_reward}, when our model is trained with GenEval alone, it often converges to shortcut solutions with simplified composition and missing background content, even when the target objects are correctly generated. Extending our framework to a multi-reward setting provides an effective counterbalance: by jointly optimizing GenEval with an auxiliary human-preference reward such as HPS, the model retains the desired objects while producing richer scenes, better aesthetics, and more natural compositions. In this sense, multi-reward training reduces the severity of reward hacking and delays its onset, although it does not eliminate the problem entirely; with sufficiently long training, shortcut behaviors can still dominate, so early stopping remains an important practical safeguard. 

Motivated by these observations, we optimize the policy against a set of reward functions $\{r^{(m)}\}_{m=1}^{M}$ using a shared latent backbone together with one lightweight value head per reward. A naive shared-head design consistently yields slow PPO progress, since different rewards have different scales and emphasize different visual attributes. Instead, each head predicts its own value $V^{(m)}_\phi(z_t,t,y)$ from the same timestep-aware latent features, while learning reward-specific scale and bias. This design amortizes the expensive diffusion feature extraction across objectives, reduces interference between heterogeneous rewards, and enables stable joint optimization in the multi-reward regime.

%% file: section/exp.tex
\section{Experiments}
\label{sec:exp}

We evaluate our post-training method on a diverse set of  image generation tasks and compare it with recent diffusion RL baselines, including DDPO and GRPO, with a focus on single-reward optimization efficiency. We then move to the multi-reward setting to test whether our shared backbone and per-reward heads can handle heterogeneous scales and signals. In addition, we present the results of our inference-time critic guidance, showing that the same critic used for PPO can be reused at sampling time to further improve the generated images without training. All experiments are run on two backbone families: (1) a UNet-based latent diffusion model, \textbf{SD 1.5}, and (2) a diffusion-transformer model, \textbf{SD 3.5 M}, to verify that our design works for both convolutional and DiT-style architectures. Finally, we provide extensive ablations on our proposed designs.

\noindent\textbf{Implementation.}
All models are trained for 300 iterations. For DDPO~\cite{black2024training} and GRPO~\cite{liu2025flowgrpo, xue2025dancegrpo} we follow the training hyperparameters in DanceGRPO~\cite{xue2025dancegrpo}. GRPO requires sampling multiple images for the same prompt to form a group and compute relative advantages, while DDPO and our PPO-based method sample only one image per prompt. To keep the comparison fair, we increase the number of distinct prompts in each iteration for DDPO and for our method so that the total number of generated images per iteration matches that of GRPO. 
Besides, we find online RL can work without classifier-free guidance~\cite{ho2020denoising}. Therefore, we disable cfg for all experiments.

\subsection{Single-Reward Benchmarks}
We first evaluate our method in the single-reward setting, where the goal is to make the model improve as fast as possible under one target reward. We group the reward models into three categories. (i) \textbf{Text-image alignment}: We use CLIP to measure how well the generated image matches the text prompt. (ii) \textbf{Human-preference rewards}: we use HPSv2.1 and PickScore, both trained on human preference data, to assign higher scores to images that are  better aligned with human judgments. (iii) \textbf{Verifiable rewards}: we use GenEval, which checks whether the compositional constraints in the prompt are satisfied, and an OCR-based text-rendering reward, which verifies whether the rendered text in the image matches the prompt.

\begin{wrapfigure}{r}{0.6\textwidth}
\vspace{-5mm}
    \centering
     \includegraphics[width=0.6\textwidth]{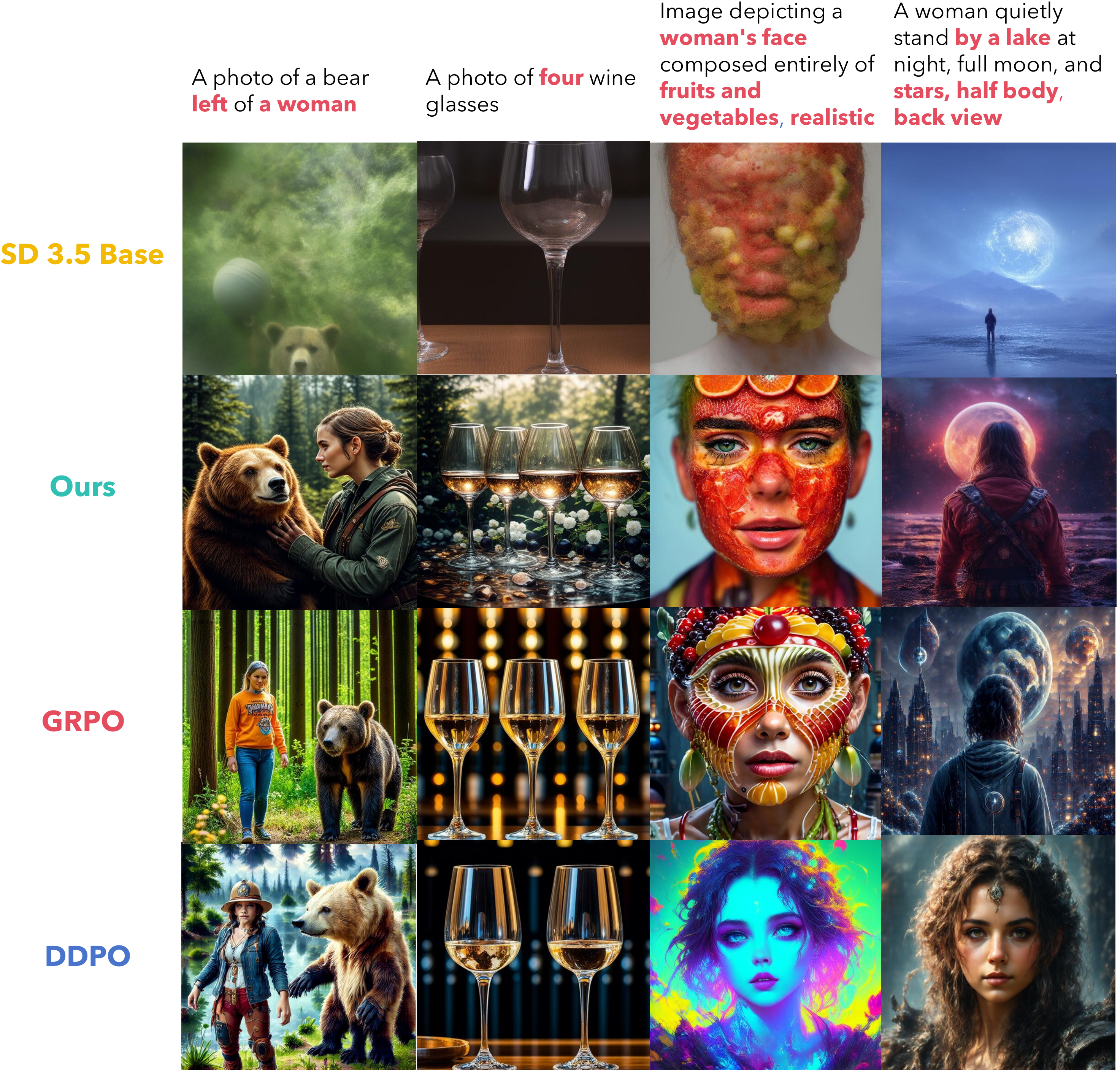}
    \caption{\textbf{Qualitative comparison between our model and baselines.}  }
    \label{fig:compare}
\vspace{-3mm}
\end{wrapfigure}

\noindent We provide the results in Table ~\ref{tab:single-reward}. Overall, PPO wins on almost all metrics for both backbones, with especially clear gains on verifiable rewards such as GenEval and OCR, suggesting that per-timestep value estimates are particularly beneficial when the reward signal is sparse and discrete. We also observe that DDPO can fail to train in some of the large-scale settings, while both GRPO and our method remain stable. GRPO needs multi-sample groups per prompt to estimate advantages, but our method, like DDPO, uses only one sample per prompt; the difference between our method and DDPO is that our advantages come from the critic trained in Sec.~\ref{sec:method}, showing that the critic can supply sufficiently accurate value estimates. Qualitative comparison in Fig.~\ref{fig:compare} shows that our method achieves better visual appearance and precise control.

\subsection{Multi-Reward Joint Optimization}
We next evaluate whether our method remains effective when multiple heterogeneous rewards must be optimized at the same time. From the three categories we pick one representative reward each: CLIP for text-image alignment, HPSv2.1 for human-preference signals, and GenEval for non-differentiable compositional constraints. We train with equal weights $1{:}1{:}1$ on these three rewards. In every iteration we also sample prompts and the corresponding metadata from the three datasets with a fixed ratio $1{:}1{:}1$, so that each update sees a balanced mix of alignment, preference, and discrete objectives.

\begin{table*}[t]
\centering
\begin{minipage}[t]{0.48\textwidth}
    \centering
    \caption{\textbf{Single-reward optimization} on SD1.5 and SD3.5-M under different reward types. Our method achieves the best results across both diffusion models and reward types. DDPO fails with SD3.5M on GenEval and OCR. }
    \resizebox{\textwidth}{!}{%
    \begin{tabular}{llccccc}
        \toprule
        Model & Method & CLIP $\uparrow$ & HPSv2.1$\uparrow$  & PickScore$\uparrow$ & GenEval $\uparrow$& OCR $\uparrow$\\
        \midrule
        \multirow{4}{*}{SD1.5}
            & Base       &  0.2250 & 0.1727  & -1.0929  &  0.1558 & 0.066  \\
            & DDPO      & 0.3233 & 0.2752  &  0.2187 & 0.5455  &  0.2351   \\
            & GRPO       & 0.3342  & 0.3779  &  0.2406 & 0.5888  &  0.2518 \\
            & Ours  &  \textbf{0.3431} & \textbf{0.3912}  &  \textbf{0.2458} & \textbf{0.6392}  &  \textbf{0.2648} \\
        \midrule
        \multirow{4}{*}{SD3.5-M}
            & Base    &  0.2275 &  0.1750 & -1.4298  & 0.2013  & 0.0002  \\  
            & DDPO     & 0.2822  & 0.2246  &  0.1976 &  div  & div  \\
            & GRPO       & 0.3006  & 0.3604  &  0.2304 & 0.9620  & 0.5653  \\
            & Ours & \textbf{0.3029}  & \textbf{0.3642}  &  \textbf{0.2320} &  \textbf{0.9758}  &  \textbf{0.5876} \\
        \bottomrule
    \end{tabular}%
    }
    \label{tab:single-reward}
\end{minipage}
\hspace{0.02\textwidth}
\begin{minipage}[t]{0.48\textwidth}
    \centering
    \footnotesize 
    \caption{\textbf{Multi-reward optimization} on SD1.5 and SD3.5-M. Our approach is the most efficient among all methods. On GenEval, our method outperforms GRPO by a large margin while remaining comparable on CLIP and HPSv2.1. }
    \label{tab:multi-reward}
    \resizebox{0.9\textwidth}{!}{%
    \begin{tabular}{llcccc}
        \toprule
        \footnotesize
        Model & Method & CLIP $\uparrow$ & HPSv2.1 $\uparrow$ & GenEval $\uparrow$ & Sum $\uparrow$ \\
        \midrule
        \multirow{4}{*}{SD1.5}
            & Base       & 0.2250 & 0.1727   &  0.1558   & 0.5535 \\
            & DDPO       & 0.2590  &  0.2995 &  0.3830 & 0.9415 \\
            & GRPO       &  0.2811 &   \textbf{0.3474}  & 0.4688 & 1.0973  \\
            & Ours &  \textbf{0.2896} &  0.3441 &  \textbf{0.4986} & \textbf{1.1323} \\
        \midrule
        \multirow{4}{*}{SD3.5-M}
            & Base     & 0.2275 &  0.1750 & 0.2013 & 0.6038   \\
            & DDPO     &  div & div  & div  & div  \\
            & GRPO       &  \textbf{0.2803} & \textbf{0.3392}  & 0.7705 & 1.3927 \\
            & Ours  &  0.2772 &  0.3130 & \textbf{0.9241}  & \textbf{1.5143}\\
        \bottomrule
    \end{tabular}%
    }
\end{minipage}
\end{table*}

The results are shown in Table ~\ref{tab:multi-reward}. As shown in the table, when optimizing multiple rewards simultaneously, all methods improve more slowly on any single metric, which is expected because the three signals can be partially conflicting. DDPO does not converge well on SD3.5-M in this mixed setting. Both GRPO and our method make stable progress on all three rewards; however, on GenEval our method is noticeably stronger than GRPO, while on CLIP and HPSv2.1 we stay close to or slightly better than GRPO. This indicates that the learned critic can capture the advantages coming from different reward sources and provide a cleaner training signal, making it more effective in the multi-reward regime. 

\subsection{Inference-Time Steering}
We further test whether the critic trained together with PPO can be used purely at inference time to steer sampling and improve rewards, without any extra finetuning. Starting from the post-trained models (SD1.5 and SD3.5-M), we compare four inference-time strategies under the same sampling budget per prompt:

(1)~\textbf{Best-of-$n$ (BoN):} draw $n$ i.i.d.\ samples and keep the highest-reward one.
(2)~\textbf{SMC:} following FK-steering~\cite{wu2023practical, zhang2025inference, singhal2025a}, maintain $n$ particles, evaluate the reward at intermediate timesteps by pushing samples to $x_0$, and resample according to normalized rewards.
(3)~\textbf{Our Critic Gradient Guidance:} add the critic-derived direction at every denoising step, generating only one sample per prompt.
(4)~\textbf{Critic Gradient Guidance + BoN:} run critic-guided sampling to generate $n$ candidates, then apply BoN selection.

\begin{wrapfigure}{r}{0.42\textwidth}
\vspace{-8mm}
    \centering
     \includegraphics[width=0.41\textwidth]{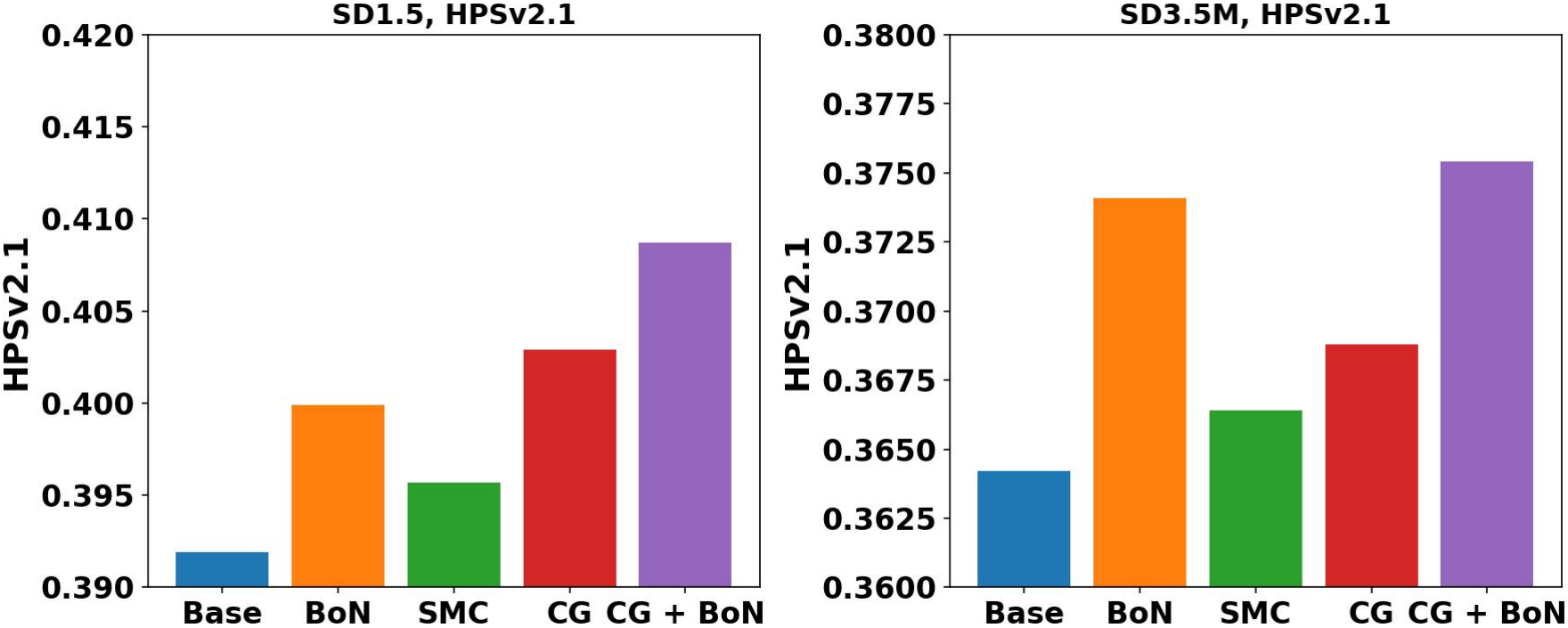}
    \caption{\textbf{Inference-time steering} on HPSv2.1 for post-trained SD1.5 / SD3.5-M. We compare BoN, SMC, our critic gradient guidance, and guidance + BoN. Base refers to the RL post-trained model.}
    \label{fig:inference-steering}
\vspace{-3mm}
\end{wrapfigure}

\noindent We evaluate on HPSv2.1 (additional rewards are reported in Appendix D).  The results are shown in Fig.~\ref{fig:inference-steering}. On SD1.5, critic guidance alone already surpasses BoN. On SD3.5-M it falls below BoN but still clearly improves over the post-trained baseline and SMC. Combining critic guidance with BoN yields the best performance on both backbones, indicating that trajectory steering and reward-based selection are complementary. Notably, critic guidance generates only a single sample per prompt, whereas BoN and SMC require $n$ forward passes—yet it matches or exceeds BoN on SD1.5 despite this $n\times$ computational advantage.

\subsection{Ablation Studies}

We conduct a series of ablations to demonstrate the effectiveness of our designs proposed in Sec.\ref{sec:method}.

\subsubsection{Critic Architecture Choices}
We first compare our diffusion model critic with several commonly used reward/vlm repurposed as a critic. All ablations are run on the HPSv2.1 reward to keep the setting consistent. We test: (i) CLIP , (ii) BLIP (iii) HPSv2.1 reward model. We then compare them to our proposed critic design.
\begin{table*}[t]
\caption{\textbf{Comparison between our proposed critic and previous pixel critics.}}
\label{tab:critic}
\centering
\begin{subtable}[t]{0.47\textwidth}
    \centering
    \caption{Ablation on critic design (HPSv2.1 reward).}
    \label{tab:ablation-critic}
    \resizebox{0.95\textwidth}{!}{
    \begin{tabular}{lcccc}
        \toprule
        Critic  &  CLIP & BLIP & HPSv2.1 & Ours\\
        \midrule
        Results $\uparrow$              & 0.3323  & 0.3612 & 0.3495 & \textbf{0.3912} \\
        \bottomrule
    \end{tabular}
    }
\end{subtable}
\hspace{0.04\textwidth}
\begin{subtable}[t]{0.47\textwidth}
    \centering
    \caption{Efficiency comparison between different critics.}
    \label{tab:critic-speed}
    \resizebox{0.95\textwidth}{!}{%
    \begin{tabular}{lcccc}
        \toprule
        Critic  & Ours & CLIP~\cite{radford2021learning} & BLIP~\cite{li2022blip} & HPSv2.1~\cite{wu2023human}\\ \midrule
        Value Computing & 10s & 33s  &   35s       & 28s \\
        Training & 20s & 78s & 86s  & 52s \\
        \bottomrule
    \end{tabular}
    }
\end{subtable}
\end{table*}

% From Table~\ref{tab:ablation-critic} we see that directly using CLIP/BLIP or even the reward model itself as the critic leads to noticeably worse final performance, and the optimization is slower. In contrast, our critic, which stays in latent space, shares the diffusion backbone and is explicitly timestep-conditioned, is much more sample-efficient and achieves higher rewards.

From Table~\ref{tab:critic}, our latent critic achieves the highest final reward on HPSv2.1, outperforming CLIP, BLIP, and using the reward model itself as a critic. Table~\ref{tab:critic-speed} further shows that pixel-space critics are substantially more expensive: our value computation takes $10$s, compared with $28$--$35$s for HPSv2.1/CLIP/BLIP (i.e., $2.8\times$--$3.5\times$ slower), and our training step takes $20$s, compared with $52$--$86$s for HPSv2.1/CLIP/BLIP (i.e., $2.6\times$--$4.3\times$ slower). Overall, operating directly in latent space with a timestep-conditioned critic yields both higher rewards and markedly better efficiency.

\begin{table*}[t]
\centering
\begin{minipage}[t]{0.44\textwidth}
    \centering
    \footnotesize
    \caption{\textbf{Ablation on value pretraining.} Value pretraining improves performance, especially on verifiable rewards.}
    \label{tab:ablation-vp}
    \resizebox{1.1\linewidth}{!}{
    \begin{tabular}{lcccc}
        \toprule
        & vp 15it & vp 50it & vp 100it & w/o vp\\
        \midrule
        GenEval $\uparrow$ & \textbf{0.6392}  & 0.6138 & 0.5932 & 0.5419 \\
        \bottomrule
    \end{tabular}
    }
\end{minipage}
\hspace{0.02\textwidth}
\begin{minipage}[t]{0.5\textwidth}
    \centering
    \footnotesize
    \caption{\textbf{Ablation on separate value heads.} Separate heads consistently outperform a shared head across all rewards.}
    \label{tab:ablation-head}
    \resizebox{0.8\linewidth}{!}{
    \begin{tabular}{lccc}
        \toprule
        & CLIP $\uparrow$   & HPSv2.1 $\uparrow$ & GenEval $\uparrow$ \\
        \midrule
        Separate head & \textbf{0.2896}  & \textbf{0.3441} & \textbf{0.4986} \\
        Shared head & 0.2804 & 0.3138 & 0.4238 \\
        \bottomrule
    \end{tabular}
    }
\end{minipage}
\end{table*}
\subsubsection{Effect of Value Pretraining}
Next, we evaluate the effect of the value-pretraining phase. We run this ablation on GenEval, which gives a sparser and less smooth signal than HPSv2.1. We compare two training settings: (i) training PPO directly from the diffusion-initialized critic without value pretraining, and (ii) doing a short value-pretraining stage before enabling actor updates. We keep all other hyperparameters the same. As shown in Table~\ref{tab:ablation-vp}, adding value pretraining improves the final performance. This supports our claim that warming up the critic gives the policy better advantage estimates, especially for hard or discrete rewards.

\subsubsection{Separate Heads for Multi-reward Training}
We study the critic design for the multi-reward setting, where we share the diffusion backbone but predict each reward with a separate lightweight MLP value head. We compare this design to a shared-head variant that regresses all rewards with a single scalar head. As shown in Table~\ref{tab:ablation-head}, separate heads consistently outperform the shared head across all rewards (CLIP, HPSv2.1, and GenEval), suggesting that decoupling value prediction reduces interference between heterogeneous objectives and better accommodates reward-specific scales and semantics.

% \begin{figure}[t]
%     \centering
%     % \includegraphics[width=0.7\textwidth]{fig/ablation_multi_head.pdf}
%     \caption{Multi-reward ablation: shared head vs.\ per-reward heads. With a single shared head the model struggles to fit rewards of different scale and noise level; per-reward heads learn faster and reach higher scores. (Figure to be inserted.)}
%     \label{fig:ablation-multi-head}
% \end{figure}
\subsubsection{Effect of CFG on training.}

\begin{wrapfigure}{r}{0.6\textwidth}
\vspace{-5mm}
    \centering
     \includegraphics[width=0.51\textwidth]{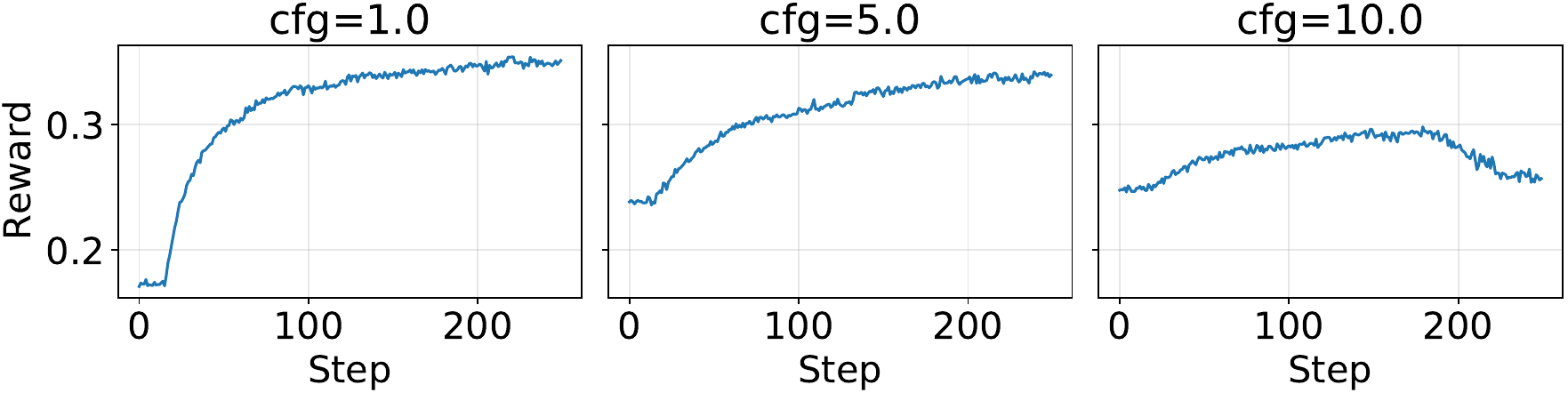}
    \caption{\textbf{Effect of CFG on training.} Higher CFG improves the initial reward, but leads to slower and less stable optimization. CFG-free ($\mathrm{cfg}=1.0$) achieves the best final reward.} 
    \label{fig:cfg}
\vspace{-3mm}
\end{wrapfigure}

We ablate the effect of classifier-free guidance (CFG) on training and use \emph{CFG-free} training ($\mathrm{cfg}=1.0$) in all main experiments. Besides reducing memory usage and improving training speed, CFG-free training also yields the best optimization behavior. As shown in Fig.~\ref{fig:cfg}, larger CFG scales improve the initial reward, but lead to slower subsequent gains and more unstable reward curves. In particular, higher CFG introduces stronger fluctuations, and $\mathrm{cfg}=10.0$ even shows reward degradation at later training stages. In contrast, $\mathrm{cfg}=1.0$ achieves both the most stable trajectory and the highest final reward.

\subsubsection{MLP conditioning}

\begin{wraptable}{r}{0.42\textwidth}
\vspace{-5mm}
    \centering
    \small
    \caption{\textbf{Ablation on critic-head conditioning.}}
    \label{tab:ablation-conditioning}
    \begin{tabular}{ccc}
        \toprule
        Time cond. & Text cond. & HPSv2.1 $\uparrow$ \\
        \midrule
        \ding{55} & \ding{55} &  0.3496\\
        \ding{55} & \ding{51} & 0.3524\\
        \ding{51} & \ding{55} & \textbf{0.3642} \\
        \ding{51} & \ding{51} & 0.3630\\
        \bottomrule
    \end{tabular}
\vspace{-3mm}
\end{wraptable}

Table~\ref{tab:ablation-conditioning} ablates conditioning for the critic MLP head. We find that timestep conditioning is essential: it improves HPSv2.1 from $0.3496$ to $0.3642$ without text conditioning, and from $0.3524$ to $0.3630$ with text conditioning. In contrast, text conditioning only provides a small gain when time conditioning is absent, and does not help once timestep conditioning is already included. We therefore use timestep conditioning for the critic head and omit additional text conditioning.

%% file: section/X_syppl.tex
% Appendix content (appendix declared in main.tex)
\newpage
\section{Appendix Overview}
\label{app:structure}
This appendix is organized as follows:
\begin{itemize}
    \item Section~\ref{app:limitations} discusses the limitations of our approach.
    \item Section~\ref{app:notation} summarizes the notation used throughout the paper.
    \item Section~\ref{app:ext_related} reviews additional related work.
    \item Section~\ref{app:inference} reports further results on inference-time steering.
    \item Section~\ref{app:details} provides additional implementation and methodological details.
    \item Section~\ref{app:vis_res} shows additional visual results and comparisons.
\end{itemize}
\section{Limitations}
\label{app:limitations}
Our approach still has several limitations. First, we focus exclusively on image diffusion models; extending our actor–critic framework to video or other modalities remains an open challenge. Second, although our multi–value-head design supports multi-reward optimization, the optimization efficiency for each individual reward can degrade compared to single-reward training. Third, our method relies on time discretization when sampling from the SDE, which may introduce discretization error and limit the RL efficiency. We leave these directions for future work.

\section{Notation}
\label{app:notation}
Table~\ref{tab:notation} summarizes the notation and abbreviations used in this paper.

\begin{table*}[t]
    \centering
    \small
    \setlength{\tabcolsep}{6pt}
    \begin{tabularx}{\textwidth}{@{}cY cY@{}}
        \toprule
        \textbf{Symbol} & \textbf{Meaning} & \textbf{Symbol} & \textbf{Meaning} \\
        \midrule
        $x_0$        & Clean data sample (image latent). &
        $s_k$        & State at step $k$, e.g.\ $(x_{t_k}, t_k, y)$. \\
        $x_t$        & Noisy latent at time $t$. &
        $a_k$        & Action at step $k$ (predicted denoising direction). \\
        $x_1$        & Standard normal latent in OT-path flow matching. &
        $r_k$        & Reward at step $k$ (zero for $k<T$, terminal-only at $k=T$). \\
        $t,\,t_k$    & Continuous time or discrete timestep. &
        $R_k$        & Return-to-go $\sum_{l=0}^{T-k}\gamma^{\,l} r_{k+l}$. \\
        $T$          & Number of reverse-diffusion / flow steps. &
        $A_k$        & Advantage estimate at step $k$. \\
        $y$          & Text prompt / conditioning input. &
        $V_\phi(s_k)$ & Value function (critic) at state $s_k$. \\
        $v_\theta(x_t,t)$ & Policy / flow network prediction (velocity / noise). &
        $\delta_k$   & TD residual $r_k + \gamma V_\phi(s_{k+1}) - V_\phi(s_k)$. \\
        $\hat{x}_0,\hat{x}_1$ & Reconstructions used in the stochastic flow sampler. &
        $\gamma$     & Discount factor in RL. \\
        $\Delta t$   & Step size between adjacent time steps. &
        $\lambda$    & GAE parameter (bias–variance trade-off). \\
        $\epsilon$   & Standard Gaussian noise, $\epsilon \sim \mathcal N(0,I)$. &
        $\pi_\theta(a_k\mid s_k,t_k)$ & Stochastic policy at step $k$. \\
        $\sigma_t$   & Noise scale / stochasticity at time $t$. &
        $\pi_{\text{ref}}$ & Frozen pretrained reference policy. \\
        $\alpha_t,\bar\alpha_t$ & DDPM schedule, $\bar\alpha_t=\prod_{s=1}^t\alpha_s$. &
        $r_k(\theta)$ & Policy likelihood ratio $\frac{\pi_\theta(a_k\mid s_k,t_k)}{\pi_{\theta_{\text{old}}}(a_k\mid s_k,t_k)}$ in PPO. \\
        $v(x_t,t)$    & Ground-truth OT-path velocity $x_1-x_0$. &
        $J(\theta)$   & Expected return. \\
        $\mathcal{T}_t$ & One-step sampler transition (DDIM / Euler). &
        $\tau$        & Trajectory $\{x_{t_k}\}_{k=0}^{T}$. \\
        $K$           & Number of particles in SMC. &
        $x_t^{(i)}$   & $i$-th particle in SMC at time $t$. \\
        $w_t^{(i)}$   & Normalized weight of particle $i$ in SMC. &
        $g_t$         & Critic gradient $\nabla_{x_t}V_\phi(x_t,t,y)$ for guidance. \\
        $\eta$        & Guidance scale for critic-based steering. &
        & \\
        \midrule
        \textbf{PPO}  & Proximal Policy Optimization. &
        \textbf{GAE}  & Generalized Advantage Estimation. \\
        \textbf{MDP}  & Markov Decision Process. &
        \textbf{BoN}  & Best-of-$N$ sampling baseline. \\
        \textbf{SMC}  & Sequential Monte Carlo sampling. &
        \textbf{UNet} & U-shaped convolutional backbone. \\
        \textbf{DiT}  & Diffusion Transformer backbone. &
        \textbf{VAE}  & Variational Autoencoder. \\
        \textbf{HPS}  & Human Preference Score (HPSv2.1). &
        \textbf{OCR}  & Optical Character Recognition reward. \\
        \textbf{SD1.5} & Stable Diffusion 1.5 backbone. &
        \textbf{SD3.5M} & Stable Diffusion 3.5M backbone. \\
        \bottomrule
    \end{tabularx}
    \caption{\textbf{Summary of the notation and abbreviations used in this paper.}}
    \label{tab:notation}
\end{table*}

\section{Extended Related Work}
\label{app:ext_related}

\noindent\textbf{Credit assignment in diffusion RL.}
A central challenge in diffusion RL is credit assignment along the denoising trajectory: rewards are typically observed only on the final decoded image, while optimization must assign credit to intermediate denoising steps. Existing methods address this problem in different ways. Early RL-based approaches such as DDPO and DPOK mainly optimize terminal rewards with policy gradients~\cite{black2024training,fan2023dpok}. Preference-based methods have explored denser or step-aware supervision, for example through temporal discounting~\cite{yang2024dense}, step-aware preference optimization~\cite{liang2025aesthetic}, and explicit denoised-distribution estimation that links intermediate steps to the final denoised outcome~\cite{shi2024preference}. Within the GRPO family, current work avoids explicit value learning and instead relies on relative group signals for trajectory optimization~\cite{liu2025flowgrpo,xue2025dancegrpo}. More recent follow-up work pushes this direction further with finer-grained reward assignment, including granular reward assessment in G$^2$RPO, dense step-wise rewards in DenseGRPO, and tree-structured temporal credit assignment in TreeGRPO~\cite{zhou2025g2rpo,deng2026densegrpo,ding2026treegrpo}. Our work is complementary to these approaches: instead of redesigning relative or dense reward signals, we learn an explicit critic for trajectory-level actor-critic optimization.

\noindent\textbf{Representation and state-space alignment.}
Another underexplored issue is the representation space in which alignment signals are defined. Many existing diffusion alignment pipelines evaluate decoded images with external reward models or vision-language encoders even when optimization concerns intermediate denoising states~\cite{zhao2025score, black2024training}. This creates a mismatch between the states visited by the policy and the states evaluated by the reward model or critic. Several recent works begin to address this issue more directly. SPO introduces step-aware preference modeling at each denoising step~\cite{liang2025aesthetic}; DDE explicitly connects intermediate states to the terminal denoised distribution for preference alignment~\cite{shi2024preference}; and LPO argues that diffusion backbones are naturally suited for modeling rewards directly on noisy latent states~\cite{zhang2025diffusion}. These works move toward diffusion-native representations, but they do not learn an explicit value critic for online actor-critic optimization. Our work shares the same intuition that alignment signals should be defined in the noisy latent space, but uses this representation to learn a timestep-conditioned value function so that actor and critic operate in the same state space.

\noindent\textbf{Diffusion/Flow Models.}
Diffusion-based generative models have achieved state-of-the-art performance in visual synthesis. Denoising diffusion probabilistic models (DDPMs)~\cite{ho2020denoising} formulate generation as reversing a fixed Markovian noising process and optimize a variational bound that can be interpreted as denoising score matching~\cite{song2019generative}. Score-based diffusion models further generalize this view by describing the forward and reverse processes as stochastic differential equations and learning time-dependent score functions, which yields a unified framework for diffusion and Langevin-based samplers and enables powerful conditional generation for diverse vision tasks~\cite{song2020score}. Flow-matching models directly learning continuous-time vector fields that transport a simple prior to the data distribution within the framework of continuous normalizing flows~\cite{lipman2023flow, liu2023flow}.  Recent analyses show that Gaussian flow matching and diffusion models are closely related and can be viewed as two parameterizations of the same underlying continuous-time generative process, offering complementary design choices in terms of objective, architecture, and discretization~\cite{lipman2024flowmatchingguidecode,gao2025diffusionmeetsflow}. In practice, flow-matching models have emerged as an attractive alternative for building fast, high-fidelity generative priors for modern vision applications.

\noindent\textbf{Post-training for Foundation Models.}
Foundation models are typically obtained by large-scale pre-training on generic corpora, and then adapted to downstream use via a post-training pipeline. Modern post-training stacks begin with supervised instruction tuning on curated demonstrations to teach the model task formats and high-level alignment objectives~\cite{ouyang2022training}. On top of this, reinforcement learning from human feedback (RLHF) trains a reward model from pairwise preferences and then optimizes the policy with reinforcement learning (e.g., PPO), improving helpfulness, safety, and calibration beyond pure supervised learning~\cite{bai2022training, lambert2024tulu}. To reduce dependence on expensive human labels, recent work explores reinforcement learning from AI feedback~\cite{bai2022constitutionalaiharmlessnessai} (RLAIF) and reinforcement learning with verifiable rewards (RLVR)~\cite{shao2024deepseekmath, team2025kimi}, where the reward signal is derived from automatically checkable criteria such as unit tests, program analyzers, or self-consistency checks, making it particularly suitable for code generation and long-horizon reasoning tasks. In parallel, direct preference optimization objectives such as DPO and related methods bypass explicit reward modeling and directly fit the policy to preference data while regularizing towards the pre-trained model~\cite{rafailov2023direct}. Recent reasoning-centric foundation models, including the OpenAI o1~\cite{jaech2024openai} series and DeepSeek-R1~\cite{guo2025deepseek}, instantiate multi-stage post-training pipelines that combine large-scale instruction tuning, RLHF-/RLVR-style optimization, and process-level supervision over chain-of-thought traces, yielding substantial gains on math, coding, and scientific reasoning benchmarks. Overall, instruction tuning, RLHF/RLAIF, RLVR-style verifiable-reward training, and direct preference optimization have emerged as the main families of post-training methods for aligning general-purpose foundation models with target tasks and user preferences.

\noindent\textbf{RL with Critics for LLM Post-training}
A large fraction of RL post-training for language models follows a critic-based design, where a value function is learned jointly with the policy to handle sparse, sequence-level feedback. In RLHF pipelines such as InstructGPT~\cite{ouyang2022training} and Constitutional AI~\cite{bai2022constitutionalaiharmlessnessai}, a reward model is first trained from human or AI preferences, and PPO with a value head on the LM is then used to maximize the reward under KL regularization~\cite{zhu2025vrporethinkingvaluemodeling}. Similar architectures appear in RL with verifiable or tool-based rewards, where the critic must propagate terminal signals back through long action sequences~\cite{liu2025asymmetric, fan2025truncatedproximalpolicyoptimization}. Recent work shows that careful choices of value normalization, credit assignment, and trajectory filtering allow PPO-style critics to scale to harder reasoning and long chain-of-thought settings~\cite{yuan2024self,seed2025thiking, guo2025seedvl, zhang2025policy}. These results suggest that strong, well-trained critics remain central when we need high performance on complex generation tasks; our work follows this line but adapts the value architecture and training to the latent, multi-step structure of diffusion models.

\noindent\textbf{Policy Gradient Methods.}
Policy-gradient methods optimize a parameterized policy by directly ascending the gradient of the expected return, rather than relying on value-based bootstrapping. Early work such as REINFORCE~\cite{williams1992simple} established the likelihood-ratio estimator and showed that subtracting a learned baseline can substantially reduce gradient variance. Actor-critic architectures extend this idea by coupling a policy (actor) with a learned value function (critic), enabling more sample-efficient and stable learning in both continuous and discrete control~\cite{konda2003onactor,mnih2016asynchronous}. Building on the natural policy gradient~\cite{kakade2001natural}, trust-region methods such as TRPO~\cite{schulman2015trust} constrain policy updates within a KL-divergence trust region to prevent catastrophic performance collapse, while generalized advantage estimation further improves the bias--variance trade-off of the policy-gradient estimator~\cite{schulman2016gae}. Proximal Policy Optimization (PPO) simplifies TRPO by replacing hard trust-region constraints with clipped surrogate objectives, yielding a robust and computationally efficient on-policy algorithm that has become the de facto choice in many large-scale RL applications~\cite{schulman2017proximal}. In parallel, deterministic policy-gradient methods and their deep variants (e.g., DDPG) extend policy-gradient ideas to off-policy learning in continuous action spaces via critic-based Q-function estimation and target networks~\cite{silver2014dpg,lillicrap2015continuous}. These developments form the algorithmic backbone for modern reinforcement-learning-based post-training of foundation models, where PPO-style policy gradients, variance-reduced advantage estimation, and critic-based shaping are widely used to optimize reward models derived from human or verifiable feedback.

\section{More Results on Inference Time Steering}
\label{app:inference}
\begin{figure}[h]
    \centering
    \includegraphics[width=0.5\textwidth]{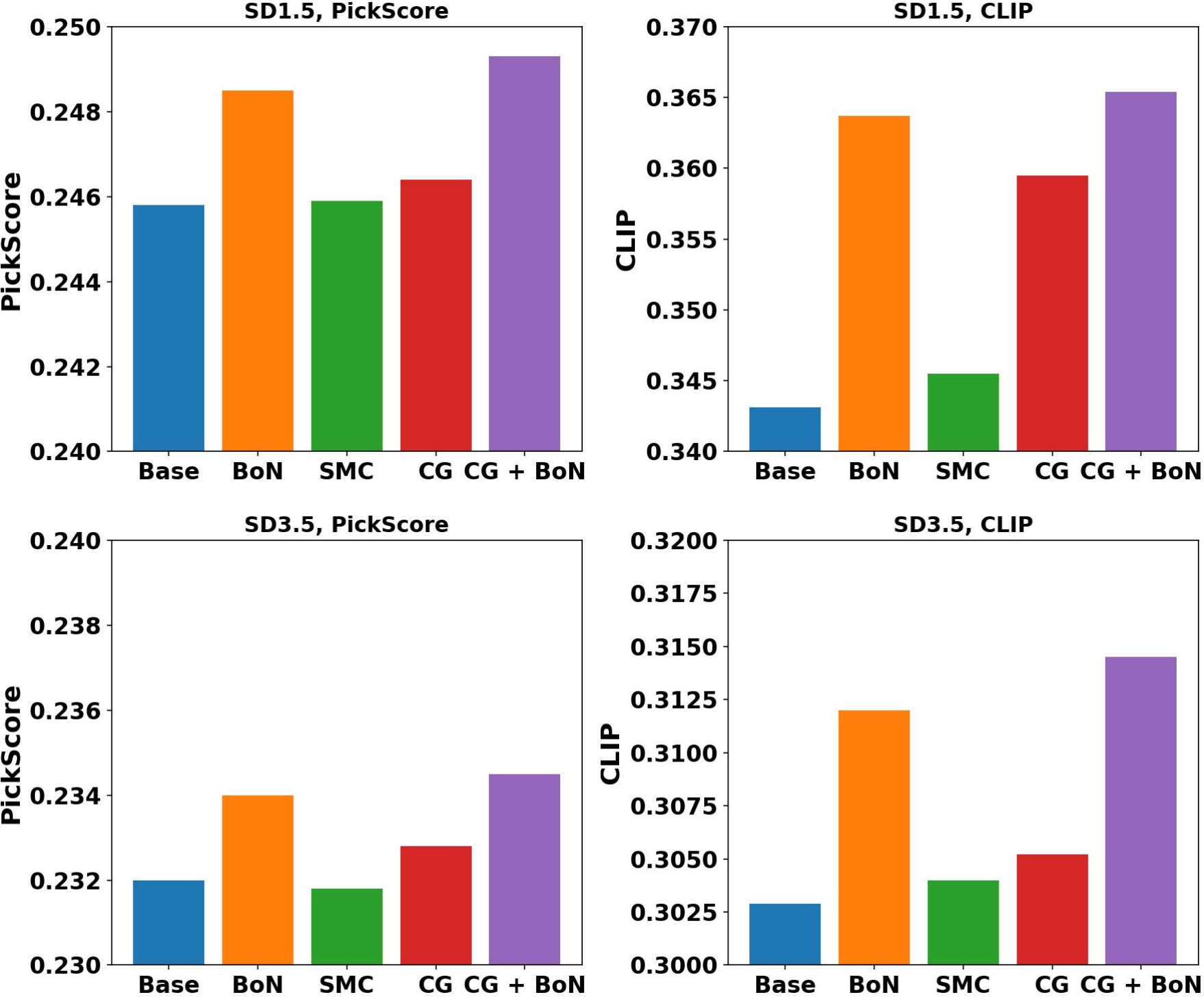}
    \caption{Inference steering results on CLIP and PickScore.}
    \label{app:fig:inference}
\end{figure}

In this section, we provide more inference time steering results on CLIP and PickScore. The results are shown in Fig.~\ref{app:fig:inference}. We can observe that critic guidance can consistently improve on baseline and BoN.
\section{Details}
\label{app:details}

\subsection{Training Details}

\noindent\textbf{Optimization.}
We use AdamW~\cite{loshchilov2018decoupled} to optimize both the policy and value networks in all experiments.
The momentum coefficients are set to $(\beta_1, \beta_2) = (0.9, 0.95)$ and the
weight decay is $1\times10^{-4}$. The policy learning rate is fixed to
$1\times10^{-4}$ across all settings for a fair comparison. Each run is trained
for 300 iterations, and every iteration consists of 4 gradient update steps. 

\vspace{2mm}
\noindent\textbf{SD1.5.}
For SD1.5 experiments, we draw a large batch of 3,840 images per iteration.
Given 4 update steps, the effective batch size per gradient update is
$3840 / 4 = 960$ samples. The PPO ratio clipping range is set to $1\times10^{-5}$.
We clip the advantages to the range $[-10, 10]$ and the value targets to
$[-5, 5]$. The learning rate of the value network is $1\times10^{-4}$. Experiments on SD1.5 require 8 H100 GPUs.

\vspace{2mm}
\noindent\textbf{SD3.5M.}
For SD3.5M experiments, we sample 384 images per iteration for the
human-preference reward and the CLIP-based reward. For other
non-differentiable rewards, we use a larger batch of 3,840 samples.
Empirically, we find that small batch sizes lead to unstable training for
DDPO/GRPO-style updates, hence the larger batch is beneficial in this case.
The ratio clipping range is set to $1\times10^{-4}$. We keep the same
advantage clipping range $[-10, 10]$ and value clipping range $[-5, 5]$.
The value network learning rate is set to $1\times10^{-3}$. Experiments on SD1.5 require 32 H100 GPUs.

\vspace{2mm}
\noindent\textbf{Sampling.}
During training, the sampling configuration is aligned with that used at
inference time. For SD1.5, we use the DDIM~\cite{song2021denoising} sampler with 50 diffusion steps.
For SD3.5M, we use the Discrete Euler sampler with 40 steps. Unless otherwise
specified, classifier-free guidance is disabled in all experiments.

\vspace{2mm}
\noindent\textbf{Normalization.}
After computing the per-sample advantages, we apply advantage
normalization over the global batch. Concretely, we compute the mean and
standard deviation of all advantages in the batch and normalize as
\begin{align}
    A_{k} = \frac{A_{k} - A_{\text{mean}}}{A_{\text{std}}}.
\end{align}
We do not apply any additional normalization to rewards or value targets.

\vspace{2mm}
\noindent\textbf{Inference Details}
At inference time, we use the same sampling configurations as in the
training phase. For SD1.5 models, we employ DDIM with 50 sampling steps.
For SD3.5M models, we use the Discrete Euler sampler with 40 steps.
Classifier-free guidance is disabled for all reported results.

\vspace{2mm}
\noindent\textbf{Multiple Reward Experiment Details}
To support multiple reward signals, we implement a unified dataset that
manages several prompt sets corresponding to different rewards. At each
training iteration, we sample prompts from these datasets according to
the proportions specified in the configuration. Each returned data item
contains the following fields: reward type, prompt,
weight, and metadata. The metadata field is empty for
human-preference rewards. For GenEval, it encodes the required number of
objects or the target color. For OCR text-rendering rewards, it stores
the desired characters that should appear in the generated image.
By assigning different weights to different reward types, we can balance
their relative contributions to the overall learning signal.

\vspace{2mm}
\noindent\textbf{Critic Attention Layers}
\label{app:critic_attn}
We augment the diffusion transformer with critic attention layers
inserted at layers $7$, $15$, and $23$. Each critic attention layer
contains a single learnable query token that aggregates information from
the intermediate visual tokens. After passing through all critic
attention layers, we concatenate the resulting query tokens and project
them to a scalar prediction, which serves as the value estimate.

Within each critic attention layer, the visual tokens are first processed
by an RMSNorm. We then obtain the query, key, and value projections
(Q/K/V). Before feeding them into the attention module, we apply QKNorm
to Q and K to improve numerical stability. The attention module produces
a single output token corresponding to the learned query. This token is
further processed by a feed-forward network (FFN) to refine the value
representation.

\begin{figure}[t!]
    \centering
    \includegraphics[width=0.3\textwidth]{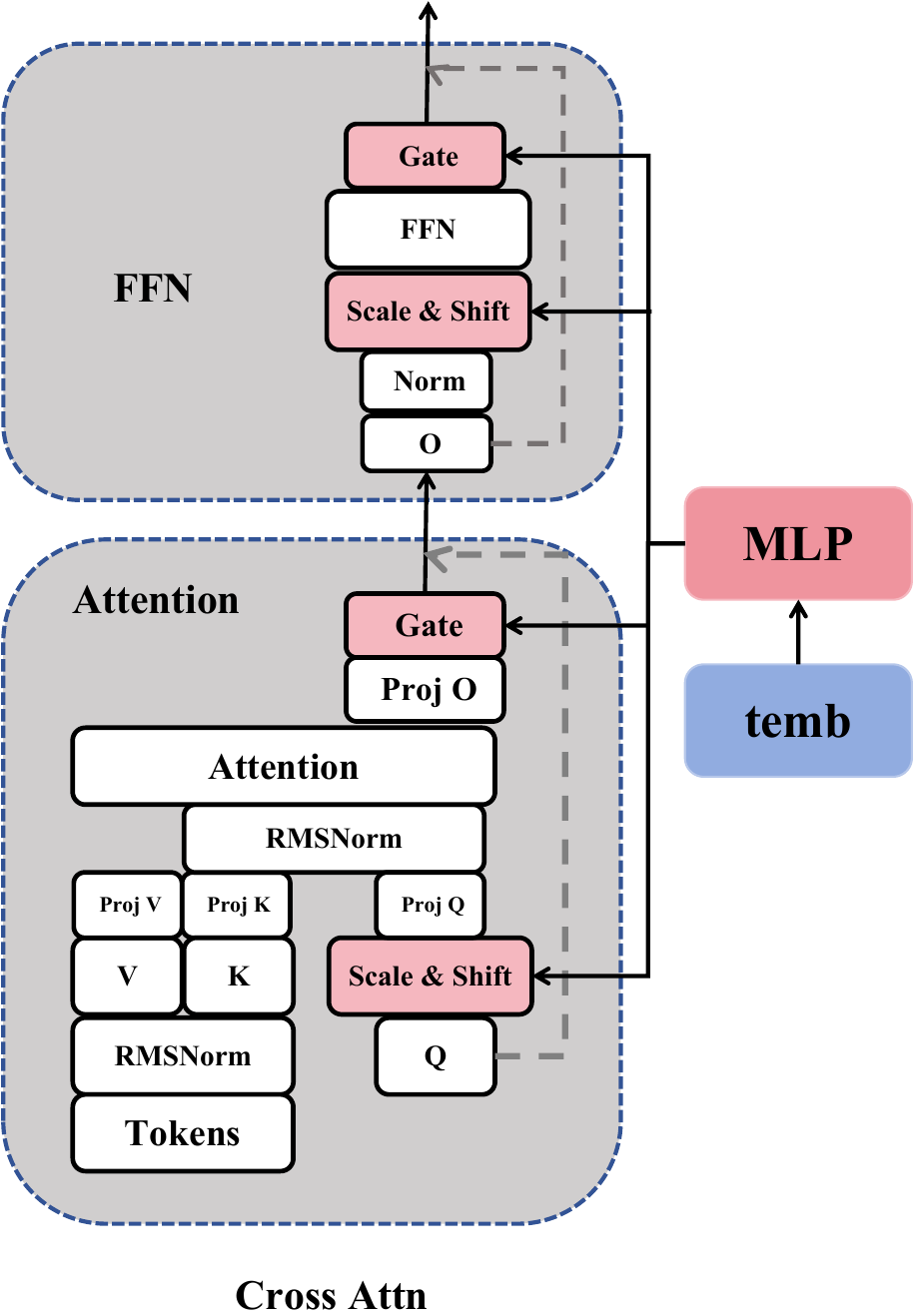}
    \caption{Architecture of the critic attention layer.}
    \label{app:fig:critic_attn}
\end{figure}

\vspace{2mm}
\noindent\textbf{Time Conditioning}
We empirically find that incorporating the diffusion timestep is crucial
for the critic. We therefore adopt an adaptive layer-normalization
(AdaLN) style time conditioning in the critic attention layers.
Specifically, the learned diffusion time embedding is first processed by
an MLP to produce per-channel \emph{scale}, \emph{shift}, and
\emph{gate} coefficients. The learnable query token is scaled and shifted
before the RMSNorm using the corresponding coefficients. After the
attention operation, the output token is gated by the residual input
using the gate coefficient. The MLP also produces scale, shift, and gate
coefficients for modulating the FFN. The overall architecture
of the critic attention layer and its time conditioning is illustrated in
Fig.~\ref{app:fig:critic_attn}.

\subsection{GAE}
We use generalized advantage estimation (GAE) to compute the advantages
for policy updates. Given a trajectory
$\{(s_t, a_t, r_t)\}_{t=0}^{T}$, we first compute the
temporal-difference (TD) residuals
\begin{align}
    \delta_t = r_t + \gamma V_\phi(s_{t+1}) - V_\phi(s_t),
\end{align}
where $V_\phi$ denotes the value network, and we set
$V_\phi(s_{T+1}) = 0$ for terminal states. The GAE advantage at time
step $t$ is then defined as
\begin{align}
    A_t^{\text{GAE}} = \sum_{l=0}^{T-t-1} (\gamma \lambda)^l \,\delta_{t+l}.
\end{align}

GAE involves two hyperparameters: the discount factor $\gamma$ and the
decay factor $\lambda$. The discount factor $\gamma \in [0, 1]$
controls how strongly future rewards are taken into account; smaller
values emphasize immediate rewards. The decay factor
$\lambda \in [0, 1]$ governs the bias--variance trade-off of the
estimator: $\lambda = 0$ corresponds to a TD(0)-style estimate with
higher bias but lower variance, whereas $\lambda = 1$ recovers the
Monte-Carlo return with lower bias but higher variance.

In our setting, trajectories are short, and the reward is provided at the
end of the trajectory. We therefore set both hyperparameters to
$\gamma = 1$ and $\lambda = 1$ in all experiments, which reduces GAE to
an undiscounted Monte-Carlo estimator of the advantages. After computing
$A_t^{\text{GAE}}$, we apply the global advantage normalization
described above.

\subsection{Reward Models}
In our experiments, we employ five reward models: CLIP, HPSv2.1, PickScore,
GenEval, and OCR. All of them take the generated image, the associated prompt, and additional metadata as input and output a scalar score, which we directly
use as the reward without extra normalization.

\vspace{2mm}
\noindent\textbf{CLIP~\cite{radford2021learning}.}
CLIP measures the text-image alignment between a generated image and its
corresponding prompt. Concretely, we encode the prompt and the image with a
pretrained CLIP text encoder and image encoder, and compute the cosine
similarity between the resulting embeddings. This similarity serves as the CLIP
reward, encouraging the model to generate images that are semantically aligned
with the input text. We use ViT-L-14 variant in our experiments.

\vspace{2mm}
\noindent\textbf{HPSv2.1~\cite{wu2023hpsv2}.}
HPSv2.1 is a human-preference score model trained on large-scale human
annotations. Given a prompt and an image, it predicts how likely a human would
prefer this image among candidates. We use the official pretrained HPSv2.1
checkpoint and feed it our generated images and prompts. The predicted score is
used as a reward that reflects human aesthetic and semantic preferences.

\vspace{2mm}
\noindent\textbf{PickScore~\cite{kirstain2023pick}.}
PickScore is another human-preference model built on top of a CLIP backbone and
trained on a large-scale dataset. For each image-prompt
pair, PickScore outputs a scalar preference score that correlates with human
judgments of overall quality. 

\vspace{2mm}
\noindent\textbf{GenEval~\cite{ghosh2023geneval}.}
GenEval is a compositional evaluation benchmark designed to test whether a
generated image satisfies fine-grained constraints such as the presence, count,
and color of objects described in the prompt. During training, we follow its
official evaluation pipeline: for each prompt, the metadata specifies the
required object categories, counts, or colors. We then apply the GenEval
detectors to the generated image and check whether all constraints are
satisfied. The GenEval reward is set to $1$ if the image passes all checks and
$0$ otherwise, encouraging the policy to generate images that strictly follow
the compositional instructions.

\vspace{2mm}
\noindent\textbf{OCR~\cite{cui2025paddleocr3}.}
To assess text rendering quality, we use an OCR-based reward. For prompts that
require specific characters or strings to appear in the image, the metadata
stores the target text. We run an off-the-shelf OCR recognizer on the generated
image to extract all visible text and compute the fraction of target
characters that are correctly recognized in the OCR output. This fraction is
used as the OCR reward, which directly encourages the model to produce images
with legible and accurate text. We use paddleocr as the reward model.

\subsection{Critic Models}

We explore several off-the-shelf vision-language models as the backbone of our
critic. In all cases, the critic takes the generated image and its
corresponding prompt as input and outputs a scalar score, which we interpret as
the value of this image-prompt pair. Concretely, we consider three families of
critics: (i) a CLIP-based critic, (ii) a BLIP-based critic, and (iii) a critic
initialized from a pretrained reward model. Our ablations are mainly conducted
on HPSv2.1, therefore we adopt the HPSv2.1-based critic as the default choice
in our experiments.

\vspace{2mm}
\noindent\textbf{CLIP Critic.}
For the CLIP critic, we use a pretrained CLIP model and keep both the image and
text encoders frozen. Given a prompt and a generated image, we obtain the text
embedding and image embedding from CLIP and compute their cosine similarity.
. This design leverages the strong
text-image alignment capability of CLIP while keeping the critic architecture
simple and stable.

\vspace{2mm}
\noindent\textbf{BLIP~\cite{li2022blip} Critic.}
BLIP is a vision-language model originally designed for image captioning and
vision-language understanding. Similar to the CLIP critic, we use the BLIP
image encoder and text encoder to extract image and text embeddings for each
image-prompt pair. We then compute the cosine similarity between these two
embeddings and take it as the BLIP critic score. Compared to
CLIP, BLIP provides a different inductive bias focused on captioning and
language understanding, offering a complementary critic signal. In our experiments, we use blip-itm-base-coco to initialize the critic model.

\vspace{2mm}
\noindent\textbf{Reward-model-initialized Critic.}
Finally, we investigate using a pretrained reward model as the initialization
for our critic. In this setting, we start from the HPSv2.1 checkpoint, which is
trained on large-scale human preference data. We use the output reward score directly as the critic
value. During our RL training, this HPSv2.1-based critic can be further
fine-tuned together with the policy, allowing the critic to stay aligned with
human preference while adapting to the distribution induced by our policy.

\subsection{Inference Guidance}
We compare our method against two inference-time baselines that only use the
frozen generator and a pretrained reward/critic model at test time:
Best-of-$N$ (BoN) and Sequential Monte Carlo (SMC). Both
Baselines operate on top of the same diffusion sampler and use the same
reward model or critic as our method to score candidate images.

\vspace{2mm}
\noindent\textbf{Best-of-$N$ (BoN).}
Best-of-$N$ is a widely-used test-time selection baseline. For each input
prompt, we independently draw $N$ samples from the diffusion model using the
standard sampling configuration described above. We then evaluate each
generated image with the corresponding reward model (e.g., CLIP, HPSv2.1, or
other task-specific reward) and obtain a scalar score. The final output for
this prompt is chosen as the image with the highest reward among the $N$
candidates. 

\vspace{2mm}
\noindent\textbf{Sequential Monte Carlo (SMC).}
Sequential Monte Carlo~\cite{doucet2001sequential} provides a more structured test-time procedure that
uses a population of samples (particles) and iteratively refines them guided by
the reward/critic. For each prompt $p$, we maintain $K$ particles
$\{x_t^{(i)}\}_{i=1}^K$ along the diffusion time steps $t=0,\dots,T$.
At every step, SMC performs propagation, reweighting, and
resampling so that computation is focused on high-reward regions
of the sample space.

We initialize particles from the standard Gaussian prior with uniform weights:
\begin{align}
    x_T^{(i)} &\sim \mathcal{N}(0, I), \quad
    w_T^{(i)} = \frac{1}{K}, \qquad i=1,\dots,K.
\end{align}
Then, for each diffusion step $t = T-1,\dots,0$, we proceed as follows:

\textbf{Propagation.}
Each particle is propagated by one step of the diffusion sampler
$\mathcal{T}_t$ (e.g., DDIM / Euler update) conditioned on the prompt $p$:
\begin{align}
    \tilde{x}_{t-1}^{(i)} = \mathcal{T}_t\bigl(x_{t}^{(i)},\, \epsilon_t^{(i)}\bigr),
\end{align}
where $\epsilon_t^{(i)}$ is the sampler noise and
$\tilde{x}_{t-1}^{(i)}$ denotes the propagated (pre-resampling) particle.

\textbf{Reweighting.}
We evaluate the critic (or reward model) $f_\psi$ on each propagated particle
and convert the scalar score into an unnormalized weight, for example, using an
exponential mapping with temperature $\tau$:
\begin{align}
    \tilde{w}_{t-1}^{(i)} &= \exp\!\Big(\tfrac{1}{\tau}\, f_\psi(\tilde{x}_{t-1}^{(i)}, p)\Big), \\
    w_{t-1}^{(i)} &= \frac{\tilde{w}_{t-1}^{(i)}}{\sum_{j=1}^K \tilde{w}_{t-1}^{(j)}}.
\end{align}
The normalized weights $\{w_{t}^{(i)}\}_{i=1}^K$ concentrate probability mass on
high-critic-score particles.

\textbf{Resampling.}
Finally, we resample particles according to the normalized weights:
\begin{align}
    a_{t-1}^{(i)} &\sim \mathrm{Categorical}\bigl(w_{t-1}^{(1:K)}\bigr), \\
    x_{t-1}^{(i)} &= \tilde{x}_{t-1}^{(a_{t-1}^{(i)})},
\end{align}
which duplicates high-weight particles and discards low-weight ones, yielding a
new equally weighted set $\{x_t^{(i)}\}_{i=1}^K$ for the next step.

After the final diffusion step, we obtain a set of particles
$\{x_0^{(i)}, w_0^{(i)}\}_{i=1}^K$. We use the critic to select the final
output as the image with the highest score:
\begin{align}
    i^\star &= \arg\max_{i} f_\psi(x_0^{(i)}), \\
    \hat{x} &= x_0^{(i^\star)}.
\end{align}

\section{Visual Results}
\label{app:vis_res}

\begin{figure*}[h]
    \centering
    \includegraphics[width=1.0\textwidth]{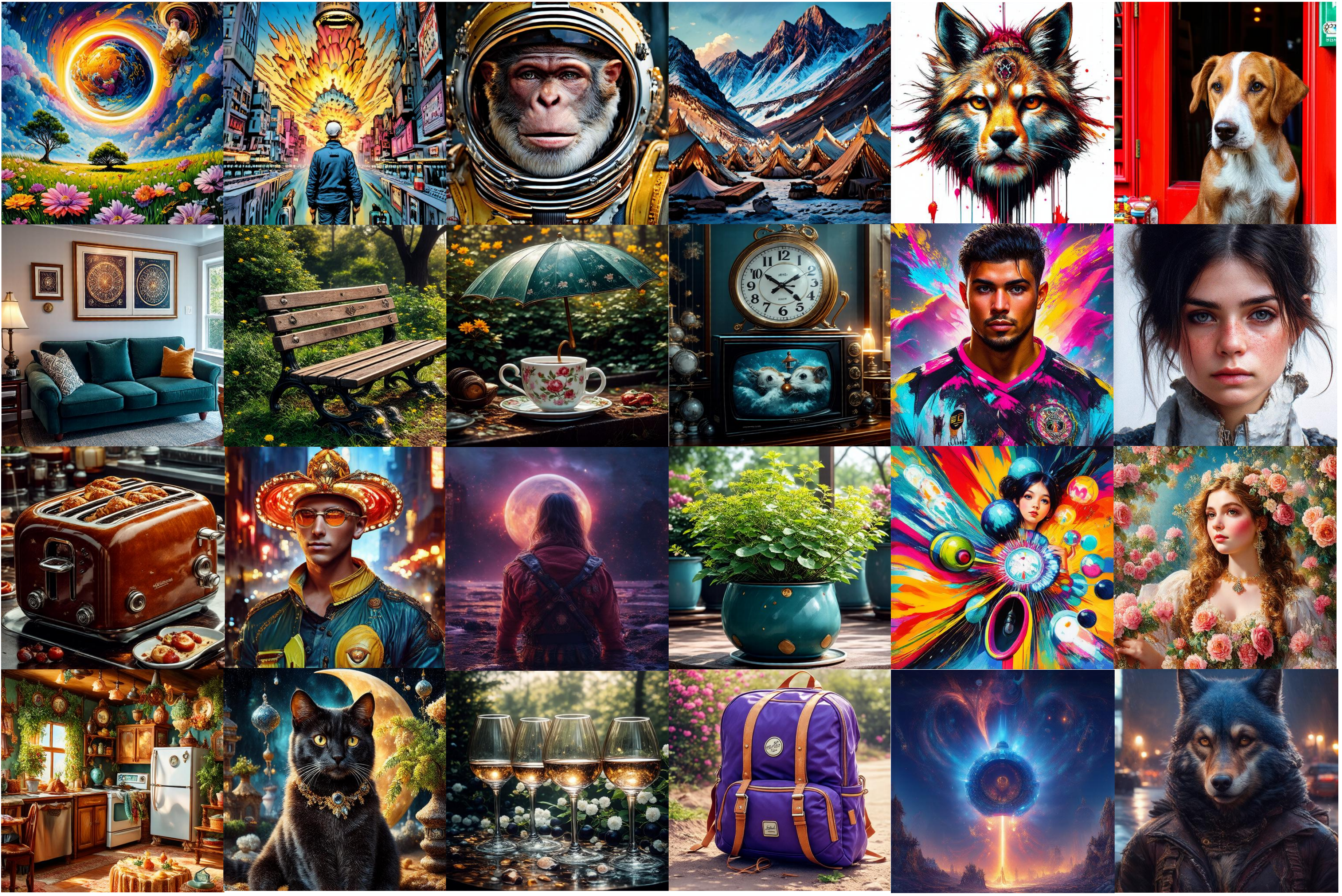}
    \caption{Additional visual results generated by our aligned models. All images are produced without classifier-free guidance (CFG).}
    \label{app:fig:results}
\end{figure*}

\paragraph{Additional Results.}
We present additional qualitative examples generated by our aligned models in Fig.~\ref{app:fig:results}.

